\documentclass[lettersize,journal]{IEEEtran}
\usepackage{amsmath,amsfonts}
\usepackage{algorithmic}
\usepackage{algorithm}
\usepackage{array}
\usepackage[caption=false,font=footnotesize,labelfont=rm,textfont=sf, captionskip=-4pt]{subfig}
\usepackage{textcomp}
\usepackage{stfloats}
\usepackage{url}
\usepackage{verbatim}
\usepackage{graphicx}
\usepackage{cite}
\usepackage{multirow}
\usepackage{enumitem}
\usepackage{booktabs}
\usepackage{soul}
\usepackage{xcolor}
\soulregister{\ref}7

\begin{document}

\title{Self-Supervised Time Series Representation Learning via Cross Reconstruction Transformer}

\author{Wenrui Zhang$^*$, Ling Yang$^*$, Shijia Geng, Shenda Hong$^\#$

\thanks{Wenrui Zhang, Ling Yang and Shenda Hong are with the National Institute of Health Data Science, Peking University, and Institute of Medical Technology, Health Science Center of Peking University, Beijing, 100191, China (e-mail: zhangwenrui@u.nus.edu, yangling0818@163.com, hongshenda@pku.edu.cn). Wenrui Zhang is also with the Department of Mathematics, National University of Singapore, Singapore, 119077, Singapore.}

\thanks{Shijia Geng is with the HeartVoice Medical Technology, Hefei, 230027, China (e-mail: gengshijia@heartvoice.com.cn).}

\thanks{$^{*}$Equal Contributions.}
\thanks{$^{\#}$Corresponding author.}

}

\markboth{Journal of \LaTeX\ Class Files,~Vol.~14, No.~8, August~2021}
{Shell \MakeLowercase{\textit{et al.}}: A Sample Article Using IEEEtran.cls for IEEE Journals}

\maketitle

\begin{abstract}
Since labeled samples are typically scarce in real-world scenarios, self-supervised representation learning in time series is critical. Existing approaches mainly employ the contrastive learning framework, which automatically learns to understand similar and dissimilar data pairs. However, they are constrained by the request for cumbersome sampling policies and prior knowledge of constructing pairs. Also, few works have focused on effectively modeling temporal-spectral correlations to improve the capacity of representations. In this paper, we propose the Cross Reconstruction Transformer (CRT) to solve the aforementioned issues. CRT achieves time series representation learning through a cross-domain dropping-reconstruction task. Specifically, we obtain the frequency domain of the time series via the Fast Fourier Transform and randomly drop certain patches in both time and frequency domains. Dropping is employed to maximally preserve the global context while masking leads to the distribution shift. Then a Transformer architecture is utilized to adequately discover the cross-domain correlations between temporal and spectral information through reconstructing data in both domains, which is called Dropped Temporal-Spectral Modeling. To discriminate the representations in global latent space, we propose Instance Discrimination Constraint to reduce the mutual information between different time series samples and sharpen the decision boundaries. Additionally, a specified curriculum learning strategy is employed to improve the robustness during the pre-training phase, which progressively increases the dropping ratio in the training process. We conduct extensive experiments to evaluate the effectiveness of the proposed method on multiple real-world datasets. Results show that CRT consistently achieves the best performance over existing methods by 2\%$\sim$9\%. The code is publicly available at https://github.com/BobZwr/Cross-Reconstruction-Transformer.
\end{abstract}


\section{Introduction}

Time series analysis \cite{9224838, 9127499, 9825701,9612727, 9669023, 9075431, 9525836, 9445591} is critical in a variety of real-world applications, such as transportation, medicine, finance, and industry. With the success of deep learning, various tasks in the field of time series analysis have achieved great performances, which include time series classification \cite{9825701,9612727}, forecasting \cite{9669023, 9075431 } and anomaly detection \cite{9525836, 9445591, chen2020imbalanced}. However, since labeled time series are usually difficult to acquire \cite{hyvarinen2016unsupervised,lan2021intrainter}, it is essential to 
study learning representations from time series data in an unsupervised learning way. Self-supervised learning is an emerging approach, which designs a pretext task and automatically generates ``labels'' for supervision while minimizing effort on annotation.

\begin{figure}
\includegraphics[width=\linewidth]{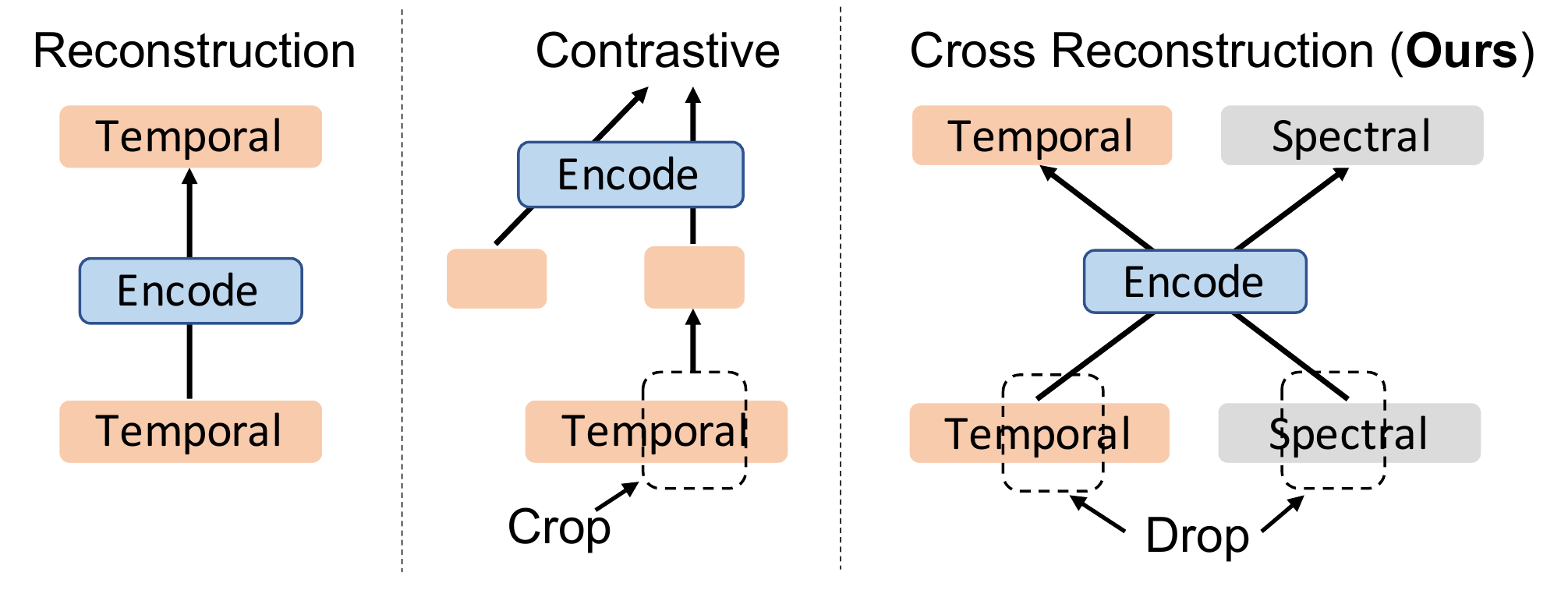}
\caption{Comparisons with previous self-supervised methods. ``Encode'' is the encoder producing the representations of time series. ``Temporal'' and ``Spectral'' are time-domain and frequency-domain data respectively. Note that our CRT utilizes both temporal and spectral information, and captures cross-domain dependencies.}\label{fig:motivation}
\end{figure}

Existing works of self-supervised learning can be grouped into two categories: contrast-based methods and reconstruction-based.
Contrast-based methods \cite{franceschi2019unsupervised,tonekaboni2021unsupervised,yang2022unsupervised} are the main-stream approach of self-supervised representation learning for time series. They are mainly to apply a segment-level sampling policy and construct positive pairs and negative pairs. The model is then forced to maximize the similarity of positive pairs, while minimizing the similarity of negative pairs in the feature space. For example, 
Contrastive Predictive Coding (CPC) \cite{oord2018representation} conducts representation learning by using powerful auto-regressive models in latent space to make predictions in the future, relying on Noise-Contrastive Estimation \cite{pmlr-v9-gutmann10a} for the loss function in similar ways. Temporal and Contextual Contrasting (TS-TCC) \cite{eldele2021time} is an improved work of CPC and learns robust representations by a harder prediction task against perturbations introduced by different timestamps and augmentations.
Reconstruction-based methods \cite{9509365, zhang2019deep} focus on utilizing the long-term context information for the time series with an encoder-decoder architecture. The pretext task of these methods is reconstructing the original data, which may be masked to enhance inference ability.
With the success of Transformer architectures \cite{9755926, 9132664, zhou2021informer}, a recent work \cite{zerveas2021transformer} proposes a Transformer-based self-supervised learning framework for multivariate time series for the first time. 

However, the existing approaches (including contrast-based and reconstruction-based learning) neglect to exploit the spectral information and utilize the temporal-spectral correlations. The frequency domain is another perspective of time series data, and different frequency bands imply various states \cite{8416667, app9071345}. Besides, each type of method is constrained by its own drawbacks: 1) segment-level sampling policy used in contrastive learning may lead to sampling bias. The performance is also unstable due to the high dependency on the way to construct negative and positive pairs. 2) the distribution shift caused by the masking process in reconstruction-based methods leads to a gap between the pre-training and fine-tuning phases. The current masking process sets parts of the time series as 0, introducing noise to the training process and destroying the shape of the time series. 

To solve these problems, we propose Cross Reconstruction Transformer (CRT) for self-supervised time series representation learning. Schematic comparisons for traditional reconstruction-based, contrast-based, and CRT methods are illustrated in Figure \ref{fig:motivation}. Our contributions include:
\begin{itemize}[leftmargin=3mm]
\itemsep0em
\item To simultaneously model the temporal-spectral information and discover correlations, we apply a Transformer encoder on both time domain and frequency domain data to automatically fuse cross-domain features. To adequately exploit the spectral information, we introduce the phase of spectral data in addition to the magnitude to store more spectral information.
\item To avoid the instability occurring in contrast-based methods, we reconstruct the original data in temporal-spectral domains based on the cross-domain representations. In addition, we propose a specified curriculum learning strategy to improve the robustness of the training process, which progressively increases the dropping ratio in the pre-training process. 
    
    \item To tackle the problem of distribution shift caused by masking in existing reconstruction-based methods, we employ a simple but reasonable ``masking'' method, which is randomly dropping certain parts of data in time and frequency domains. Instead of setting values as 0, dropping discards the ``masked'' segments and preserves the original distribution of the time series. 

    \item We evaluate the representations learned by our proposed method on three datasets, and the results demonstrate that our CRT achieves the best performance in terms of effectiveness and robustness, with a performance gain of 2\%$\sim$9\%. 
\end{itemize}

\section{Related Work}
\subsection{Self-supervised learning}
As an emerging research field, self-supervised learning has drawn much attention, especially for image data and text data \cite{denton2017unsupervised,9772739,wang2015unsupervised,pmlr-v119-chen20j,9745754}. Self-supervised learning is to manually design ``pretext tasks'' on unlabelled data, which is a way to provide ``supervised'' information for deep learning models. The model is expected to capture inherent characteristics among large-scale unlabelled data, and show a better performance on downstream tasks compared to training from scratch \cite{yang2022omni}. 

For text data, BERT \cite{devlin2018bert} is a typical example of self-supervised learning, which is designed to pre-train deep bidirectional representations from the unlabeled text by joint conditioning on both left and right context in all layers. Due to BERT's great performance, multiple variations were proposed. For example, SpanBERT proposed to mask random contiguous spans, rather than random individual tokens \cite{joshi2020spanbert}. ERNIE proposed entity-level masking and phrase-level masking, which integrate phrase and entity-level knowledge into the language representation \cite{zhang2019ernie}. For image data, contrastive learning is more popular to learn representations in an unsupervised way. Contrastive learning methods typically rely on discriminating manually constructed pairs through an InfoNCE loss \cite{oord2018representation}. InfoNCE maximizes the similarity between positive instances and minimizes the similarity between negatives. The largest challenge and difficulty is how to construct reasonable and practical positive and negative sets to facilitate the model to capture similarities and differences among instances. Some works propose to construct pairs via data augmentations methods \cite{pmlr-v119-chen20j, chen2020improved}, clustering \cite{caron2018deep, caron2020unsupervised}, and so on. Contrastive learning is also applied to other fields, such as graph representation learning \cite{9945993}, multi-modal learning \cite{zhang2021cross}, and so on. However, contrastive learning heavily depends on constructing positive and negative pairs, which makes the performance less robust and unstable. Recent works have started to apply reconstruction-based self-supervised learning on image and other data \cite{he2022masked, jiang2022dual}, which are straightforward but show great performance.

\subsection{Self-supervised learning for time series}
Research on self-supervised representation learning on sequence data has been well-studied, but representation learning for time series still needs to be promoted. Inspired by the well-studied self-supervised learning methods in computer vision and natural language processing areas, recent works mainly leverage the contrastive learning framework for time series representation learning \cite{yang2022unsupervised}. The researchers design different time-slicing strategies to construct positive and negative pairs with the assumption that temporally similar fragments could be viewed as positive samples and remote fragments are treated as negative samples \cite{yue2021ts2vec}.
Unsupervised Scalable Representation Learning \cite{franceschi2019unsupervised} introduces a novel unsupervised loss with time-based negative sampling to train a scalable encoder, shaped as a deep convolutional neural network with dilated convolutions \cite{oord2016wavenet}.
Temporal Neighborhood Coding \cite{tonekaboni2021unsupervised} introduces the concept of a temporal neighborhood with stationary properties as the distribution of similar windows in time.
Nevertheless, these time-based methods are sometimes unreasonable for long time series and fail to capture the long-term dependencies \cite{yang2022unsupervised}. Besides, the performance will substantially deteriorate when applied to downstream tasks containing periodic time series.

A recent work \cite{zerveas2021transformer} proposes a Transformer-based reconstruction self-supervised task for time series data for the first time. This method masks part of the original time series data by setting the values as zeros, and uses a linear layer to reconstruct the masked data. The results show that reconstructing the masked data can help extract dense vector representations of multivariate time series and facilitate models to understand contextual information. However, masking original data introduces a gap between the pre-training stage and fine-tuning stage, because it changes the original distribution of time series significantly. In addition, all methods including contrastive learning methods neglect an important characteristic of time series data - the frequency domain.

\begin{figure*}
    \centering
    \includegraphics[width=0.89\linewidth]{ 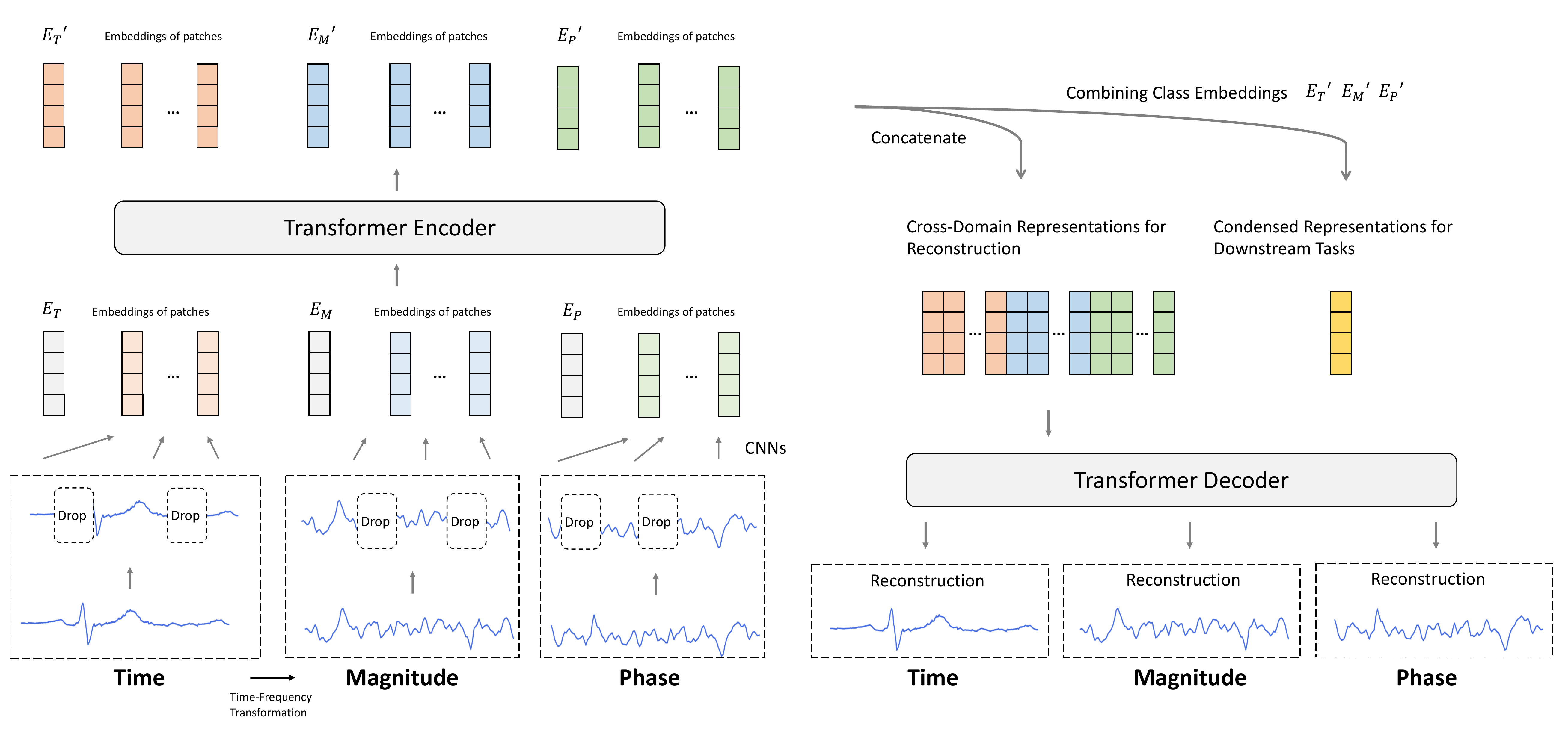}
    \caption{The framework of our CRT. For each time series, we transform it into the frequency domain and calculate the magnitude and phase as explained in Section \ref{sec:frequency}. We then slice the data and randomly drop some sliced patches. The remaining patches are projected as three types of embeddings through three convolutional neural networks (shown as arrows in the figure). Decorated with the prefixed [CLS] tokens ($E_T$, $E_M$, and $E_P$) and summed with position embedding and domain-type embedding, all embeddings are fed into a cross-domain Transformer encoder. For our pre-trained reconstruction task, we feed all resultant cross-domain representations into the decoder to reconstruct the original time, magnitude, and phase data. For downstream tasks, we condense $E_T'$, $E_M'$ and $E_P'$ (corresponding to $E_T$, $E_M$ and $E_P$) as the representation. }
    \label{fig:overview}
\end{figure*}

\section{Cross Reconstruction Transformer}
The framework of CRT is shown in Figure \ref{fig:overview}. Our CRT is a Transformer-based time-frequency domain cross-reconstruction method. The goal of our method is to learn representations combining useful information of two domains. Generally, the model is firstly pre-trained following our CRT framework, and then transferred to the downstream tasks. In this section, we describe our CRT in detail. All important notations are shown in Table \ref{tab:notations}.

At first, we pay attention to the drawbacks of existing self-supervised learning methods for time series data:

\begin{itemize}[leftmargin=5mm]
    \item The existing reconstruction methods for time series mask original data and reconstruct it. However, masking (set to zeros) can significantly change the original pattern of the time series and bring many noises to the pre-training process that would deteriorate the performance. To tackle these problems, we propose the dropping approach, which is discussed in Section \ref{sec:dropping}.
    \item Most representation learning methods for time series neglect the frequency domain, which is a complementary perspective to the time domain of time series. In Section \ref{sec:frequency}, we propose a novel way to better utilize the spectral information and model temporal-spectral correlations.
\end{itemize}
Combining the above two modules, we propose our Dropped Temporal-Spectral Modeling for time series and discuss it in Section \ref{sec:cross}. In Section \ref{sec:optimization}, we introduce an effective progressive training strategy to improve the stability and capacity of our CRT.

\subsection{Dropping Rather Than Masking}
\label{sec:dropping}

\begin{figure}[ht]
    \centering
    \includegraphics[width=\linewidth]{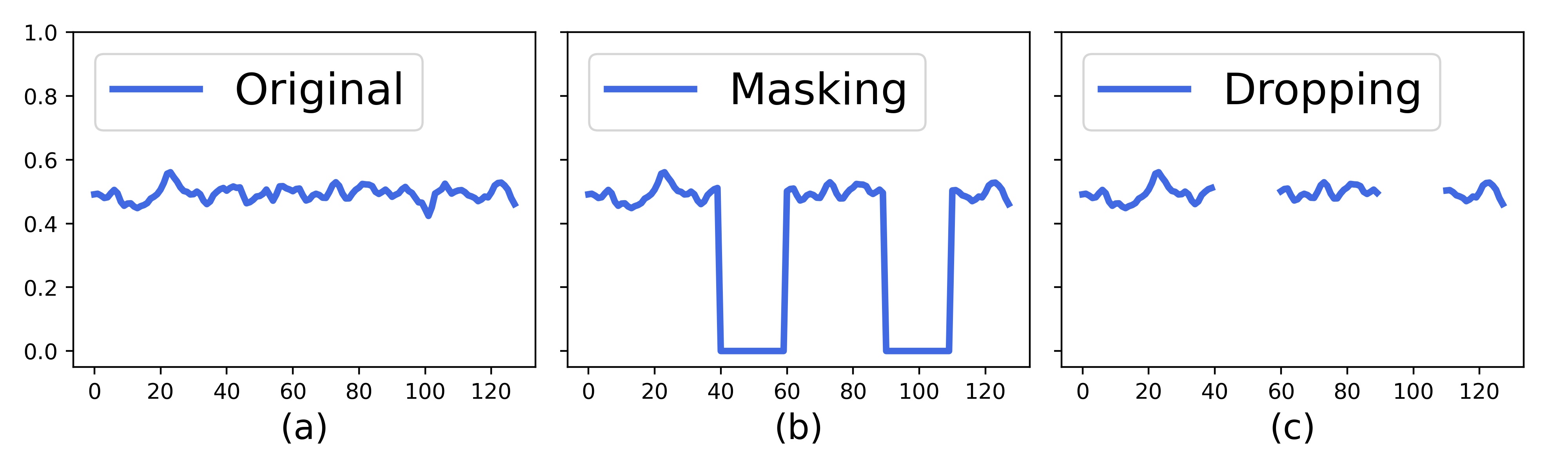}
    \caption{Masking vs. Dropping. Given one time-series data (a), masking means to set segments of values to zeros (b), and dropping means to discard some segments (c).}
    \label{fig:dropping}
\end{figure}

Masking parts of data and reconstructing them is a common paradigm in self-supervised learning tasks \cite{devlin2018bert, he2022masked}. Inspired by this, some works mask the original time series by setting the value of certain temporal segments to 0 and reconstructing these segments \cite{zerveas2021transformer}. However, this may significantly change the original pattern of time series and bring noises to the representation learning process. Besides, it may cause the gap between the pre-training stage and fine-tuning stage since it leads to distribution shifts. 
As shown in Figure \ref{fig:dropping}(b), masking significantly changes the shape and distribution of the original time series. Different from images or text, the shape is an important and unique pattern for time series \cite{ye2009time}. Thus masking would bring noises into the pre-training stage and deteriorate the downstream performance.

In this work, we choose to drop some segments of the time series rather than masking them. Dropping discards certain parts of the input time series, and feeds the rest of the data into the model. The difference between masking and dropping is shown in Figure \ref{fig:dropping}. In detail, we slice the input into patches with the same length, and randomly discard patches in a probability of $r$ following:

\begin{equation}
    Dropping = \left\{
    \begin{array}{lr}
        \text{True} & p \leq r \\
        \text{False} & p > r
    \end{array}
    \right.
\end{equation}

where $p$ is a random value following uniform distribution $U(0,1)$.
In this way, the model receives segments of real data without corrupted zeros portions. Adding corresponding position information of undropped parts to the Transformer model can maximally preserve the global context and capture the intrinsic long-term dependencies.

\subsection{Involving Spectral Information} \label{sec:frequency}
\begin{figure}
\centering
    \includegraphics[width=\linewidth]{ 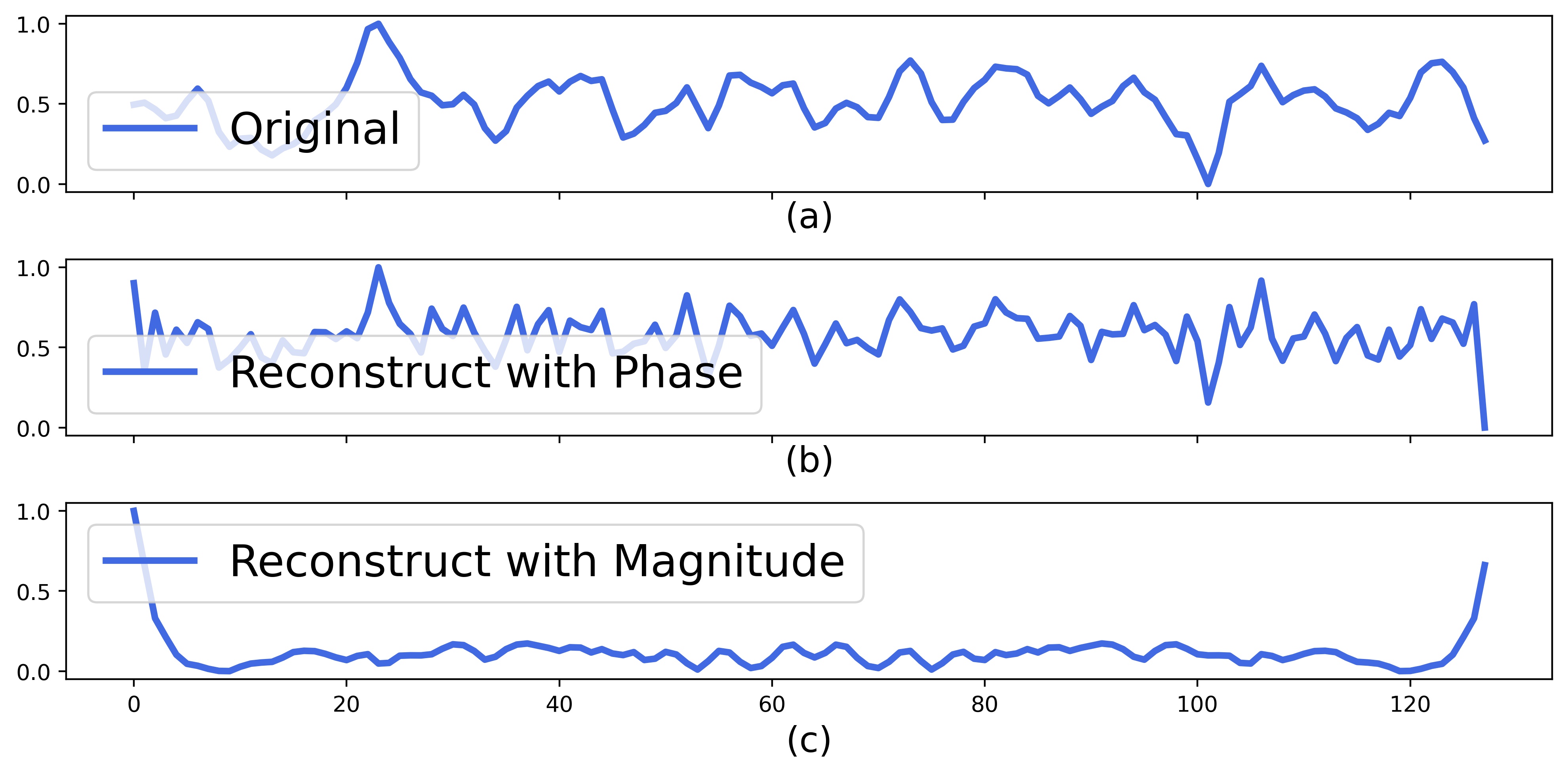}
	\caption{Phase v.s. Magnitude. Given a time series (a), reconstruction with phase tends to preserve the shape of the original data (b), while reconstruction with magnitude tends to preserve the value of the original data (c). All data are conducted min-max normalization. The Euclidean distance between (b) and (a) is 1.62, and Euclidean distance between (c) and (a) is 5.05.}\label{fig:phase_magnitude}
\end{figure}

When learning representations for time series, most of the existing methods neglect to involve the spectral information of time series. The frequency domain provides another perspective to discover the patterns for time series data. The spectral perspective can provide several characteristics of the time series that are not easily acquirable in the time domain.
Consequently, we attempt to explicitly input both time and frequency domains into the model to enable it to learn both temporal and spectral information and model cross-domain correlations. 

The most common method to transform time series into the frequency domain is the Fast Fourier Transform (FFT) \cite{Oppenheim}, which is a rapid implementation of the Discrete Fourier Transform (DFT): 
\begin{equation}
    \mathcal{F}[k]=\sum_{n=0}^{N-1}\mathbf{t}[n](\cos(\frac{2 kn \pi}{N})-\sin(\frac{2 kn \pi}{N})i), k = 0,1,\dots,N-1
\end{equation}
 where $\mathbf{t}$ is original time series data, $N$ is the length of $\mathbf{t}$, $k$ is the index of frequency data, $i$ is called imaginary unit satisfying $i^2=-1$, and $\mathcal{F}$ is the frequency domain data. After FFT, the original time series is transformed into the frequency domain, and is converted to a sequence of complex numbers. However, these complex numbers cannot be used to train the neural networks directly. Most of the methods considering the frequency domain attempt to store the spectral information using the magnitude of the complex numbers. For any complex number $z = a+bi$, where $a$ and $b$ are arbitrary real numbers. The magnitude of $z$ is $||z||=\sqrt{a^2+b^2}$. 
However, using only magnitude to restore the complex numbers would cause information loss. Because we cannot restore $z$ only using $||z||$.

To tackle this problem, we propose a novel way to better utilize the spectral information rather than simply using the magnitude. We introduce phase $\phi(z) \in (-\pi, \pi]$ as follows:
\begin{equation}
\centering
\phi(z)=
\left\{
\begin{array}{lr}
    \arctan\frac{b}{a} & a > 0\\
    \arctan\frac{b}{a} + \text{Sign}(b) \times \pi & a < 0\\ 
    \text{Sign}(b)\times \frac{\pi}{2} & a = 0
\end{array}
\right.
\end{equation}
where 
\begin{equation*}
    \text{Sign}(b) = 
\left\{
\begin{array}{lr}
    \frac{||b||}{b} & b \neq 0 \\
    0 & b = 0 
\end{array}
\right.
\end{equation*}
Phase is another characteristic of the frequency domain. Intuitively, a complex number $z$ can be regarded as a point $P$ of a circle with radius of $||z||$ and the angle between $\vec{OP}$ and the positive direction of x-axis. With phase and magnitude, we can restore the full spectral information which is represented by the complex number $z$ through

\begin{align}\label{equ:restore}
\begin{split}
    a &= ||z||\times \sin(\phi(z)),\\
    b &= ||z||\times \cos(\phi(z)),\\
    z &= a+bi
    \end{split}
\end{align}

In addition to complementing magnitude, the phase contains more information relative to the shape of the times series. We compare the information contained by phase and magnitude by restoring the original time series only using phase or magnitude. In detail, we calculate the phase and magnitude of a time series, and replace the phase or magnitude with a constant vector. In both cases, we restore the complex vector following Equation \ref{equ:restore}, and conduct inverse FFT to transform complex vectors to the time domain. As shown in Figure \ref{fig:phase_magnitude}, we can see that the original time series is more similar to the restored time series with the correct phase. This intuitively explains the importance of phase information for the shape of time series data. 

\begin{table}[]
\centering
\caption{Notations.}
\resizebox{\columnwidth}{!}{
\begin{tabular}{c|l}
\toprule
Notation & Definition \\
\midrule
$\mathbf{t}$ & The original time series \\
$\mathbf{m}$ & The magnitude of $\mathbf{t}$\\
$\mathbf{p}$ & The phase of $\mathbf{t}$\\
$\mathbf{x}$ & Concatenation of $\mathbf{t}$, $\mathbf{m}$ and $\mathbf{p}$ \\
$r$ & Dropping ratio \\
$N$ & Number of patches of $\mathbf{x}$\\
$B$ & Number of $x$ in a batch (batch size)\\
$\mathbf{x_i}$ & The i-th $\mathbf{x}$ in a batch of $\mathbf{x}$\\
$\mathbf{E_i}$ & The projection of $\mathbf{x_i}$ by convolutional neural networks\\
$\mathbf{E_i'}$ & The projection of $\mathbf{E_i}$ by Transformer\\
$\mathbf{E_T}$ & The [CLS] token for time domain\\
$\mathbf{E_T'}$ & The projection of $\mathbf{E_T}$ by Transformer\\
$\mathbf{E_M}$ & The [CLS] token for magnitude.\\
$\mathbf{E_M'}$ & The projection of $\mathbf{E_M}$ by Transformer\\
$\mathbf{E_P}$ & The [CLS] token for phase\\
$\mathbf{E_P'}$ & The projection of $\mathbf{E_P}$ by Transformer\\
\bottomrule
\end{tabular}
}
\label{tab:notations}
\end{table}

\subsection{Dropped Temporal-Spectral Modeling} \label{sec:cross}

In this section, we introduce our Dropped Temporal-Spectral Modeling. For a given time series data $\mathbf{t}$, we firstly conduct FFT and convert $\mathbf{t}$ from the time domain to the frequency domain. The obtained vector is a complex vector, of which we store both the magnitude and the phase as mentioned in Section \ref{sec:frequency}. Thus, the complex vector is transformed into two real vectors $\mathbf{m}$ (magnitude) and $\mathbf{p}$ (phase) with the same length. We then concatenate $\mathbf{t}$, $\mathbf{m}$, and $\mathbf{p}$ to get the input $\mathbf{x} = [\mathbf{t},\mathbf{m},\mathbf{p}]$ of length $L$.

For the input $\mathbf{x}$, we choose to drop parts of it as mentioned in Section \ref{sec:dropping}. Each of the sliced patches contains only one type of data (time, magnitude, or phase). After that, we drop $r \times N$ ($r \in (0, 1)$ is the given dropping ratio and $N$ is the number of patches) patches and use remaining $(1-r) \times N$ patches to reconstruct the dropped patches. For patches containing time information, we randomly drop them. Also, for patches containing frequency information (phase or magnitude), we drop the corresponding pair of patches (phase and magnitude of the same segments of complex vectors). As explained in Section \ref{sec:dropping}, dropping the patches can reduce the gap between the pre-training stage and fine-tuning stage because it does not change the shape of the original data.

Before the Transformer encoder, we add three convolutional neural networks (CNNs) to extract local features for three types of patches respectively. We hypothesize CNN can help extract local features, and Transformer can help model cross-domain information using the local features. The patches $\mathbf{x_i}$ from time domain and magnitude and phase are projected to three types of embeddings $\mathbf{E_i}$ respectively. We then add a different learnable [CLS] token before each type of embeddings (the [CLS] before the time, magnitude, and phase embeddings are referred to as $\mathbf{E_T}$, $\mathbf{E_M}$ and $\mathbf{E_P}$ respectively). The three types of embeddings are summed with their corresponding learnable domain-type embeddings and concatenated into a vector $\mathbf{E} \in \mathbb{R}^{(N+3)\times D}$, where $D$ is the dimension of the embeddings. Before fed into the Transformer, $\mathbf{E}$ is summed with the corresponding learnable position embeddings.

We input the whole $\mathbf{E}$ into the Transformer encoder and use the produced features to reconstruct the original data (time, magnitude, and phase). The learned representations are used to reconstruct the original data of both domains, so that the Transformer encoder can automatically capture complementary information from both domains.

Formally, in a batch, there are $B$ inputs $\mathbf{x_i}, i=1,\dots,B$. $\mathbf{x_i}$ is projected to $\mathbf{E_i}\in \mathbb{R}^{(N+3)\times D}$ by CNNs, and $\mathbf{E_i}$ is projected by Transformer to $\mathbf{E_i'} \in \mathbb{R}^{(N+3)\times D}$. We use $\Bar{\mathbf{E_i‘}} = \frac{\mathbf{E_T'}+\mathbf{E_M'}+\mathbf{E_P'}}{3}$ for down-stream tasks. We concatenate each $\mathbf{E_i'}$ with learnable decoding tokens, and feed them into the decoder to reconstruct all patches. The reconstruction loss is:
\begin{equation}
    \mathcal{L}_{Recon} = \frac{1}{L\times d}\sum_{i=1}^L\sum_{j=1}^d(\mathbf{x'}[i][j] - \mathbf{x}[i][j])^2
\end{equation} where $\mathbf{x'}$ is the reconstructed input (including both dropped and undropped patches), and d is the dimension of $\mathbf{x}$.

When reconstructing the original data of both domains, we hope the latent representations contain more sample-specific information and less similar information to other samples. Hence, we propose a simple Instance Discrimination Constraint (IDC) module to remove redundancy and sharpen decision boundary. IDC maximizes the mutual information between the representations and original input, and simultaneously constrains the redundant information that other samples have in common. Also, we maximize the distance between $\mathbf{\Bar{E_i'}}$ and $\mathbf{E_j}$ where $i \neq j$ in latent space, that is minimizing:

\begin{align}
    \mathcal{L}_{IDC}& = \frac{1}{B(B-1)}\log[\sum_{i=1}^B \sum_{j=1..B, j\neq i} e^{sim(Proj_1(\mathbf{\Bar{E_i'}}), Proj_2(\mathbf{E_j}))}]
\end{align} where $B$ is the batch size, $sim(\cdot,\cdot)$ is cosine similarity $sim(x,y)=\frac{x\cdot y}{||x||||y||}$, and $Proj_1, Proj_2$ are two projection heads to project $\mathbf{\Bar{E_i'}}$ and $\mathbf{E_j}$ into the same latent space. The projection heads can be either a linear layer or a multi-layer perceptron, and we use a two-layer perceptron in our experiments.

\subsection{Model Optimization} \label{sec:optimization}
Combining the reconstruction loss and IDC loss, our final loss is:
\begin{equation}
    \mathcal{L} = \mathcal{L}_{Recon} + \beta \mathcal{L}_{IDC}
\end{equation} where $\beta$ is a hyper-parameter. By updating this loss, we hope the representations learned by the model contain as much sample-specific information as possible. Thus, the representations is both informative and discriminable. As for the complexity, the FFT process is conducted before pre-training the model and frequency data is directly used during self-supervised training. Consequently, the increase in computational complexity of optimizing $\mathcal{L}$ is negligible.

However, it is difficult to reconstruct the dropped patches when $r$ is initially set as a large value. As a result, we introduce Curriculum Learning (CL) \cite{bengio2009curriculum}, which is a strategy that trains the model from ``easy'' to ``hard''. In detail, we set a relatively small dropping ratio $r$ at the beginning of the pre-training stage, and increase $r$ as the self-supervised learning goes. Formally, we refer to the initial $r$ as $r_{min}$, final $r$ as $r_{max}$, and the number of epochs for increasing $r$ as $N_{epoch}$. $r_{min}$ and $r_{max}$ are hyperparameters, and should be tuned on the validation set. Then, the $r$ for each epoch $i$ is 
\begin{equation}
    r_i = \max(r_{min}, \min(r_{max}, \frac{i}{N_{epoch}}))
\end{equation}
When the current epoch $i$ is smaller than $r_{min} \times N_{epoch}$, $r_i$ is equal to $r_{min}$. When $r_{min} \times N_{epoch} \leq i \leq r_{max} \times N_{epoch}$, $r_i$ is equal to $\frac{i}{N_{epoch}}$. Then, $r_i$ remains equal to $r_{max}$ until the pre-training stage ends.

\section{Experiments}

\subsection{Experimental Setup}

\subsubsection{Datasets}

We conduct experiments on three publicly available time series datasets:

\begin{itemize}[leftmargin=5mm]
\itemsep0em
\item \textbf{PTB-XL} \cite{wagner2020ptb,goldberger2000physiobank}.
PTB-XL is an electrocardiogram (ECG) dataset with 21,837 12-lead ECGs of 10s, where 52\% are male and 48\% are female, ranging in age from 0 to 95. The sample rate is 500 Hz, and the length of each recording is 5,000. Our task is to classify time series into one of the five classes including normal ECG (NORM), conduction disturbance (CD), hypertrophy (HYP), myocardial infarction (MI), and ST/T change (STTC). 
\item \textbf{HAR} \cite{anguita2013public}. HAR is a human activity recognition dataset with 10,299 9-variable time series with a length of 128 from 30 individuals. The data are sampled from the accelerometer and gyroscope with a sampling rate of 50 Hz. Our task is to classify time series into one of the six classes of daily activity, including walking, walking upstairs, downstairs, standing, sitting, and lying down.
\item \textbf{Sleep-EDF} \cite{kemp2000analysis,goldberger2000physiobank}. Sleep-EDF is built for sleep stage classification. We use data from 4 subjects in their normal daily life recorded by a modified cassette tape recorder. With the sampling rate of 100 Hz, we choose 3 channels: horizontal EOG, FpzCz, and PzOz EEG. Each recording is a 24-hour time series, and we cut them into 30-s segments. Our task is to classify time series into one of the eight classes including wake, sleep stage 1 - 4, rapid eye movement, movement, and unscored. 
\end{itemize}
The overview of our adopted datasets is shown in Table \ref{tab:dataset}.

All three datasets are randomly split into three parts, training set (80\%), validation set (10\%), and test set (10\%). In real-world scenarios, unlabeled data is much more common than labeled data. To better simulate this situation, we use the whole training set to pre-train, and randomly select 20\% of the training set to fine-tune the pre-trained model. 

\begin{table}[]
\centering
\caption{Statistics of datasets. }
\resizebox{\columnwidth}{!}{
\begin{tabular}{c|c|c|c|c}
\toprule
Dataset & \# of samples & \# of channels & \# of classes & Length\\
\midrule
PTB-XL & 21,837 & 12 & 5 & 5000 \\
HAR & 10,299 & 9 & 6 & 128 \\
Sleep-EDF & 22,636 & 3 & 8 & 3000 \\
\bottomrule
\end{tabular}
}
\label{tab:dataset}
\end{table}

\begin{table}
\centering
\caption{Key hyper-parameters of datasets. }
\begin{tabular}{c|c|c|c|c}
\toprule
Dataset  & patch size & $r_{min}$ & $r_{max}$ & $N_{epoch}$\\
\midrule
PTB-XL & 20 & 0.3 & 0.6 & 200 \\
HAR & 8 & 0.3 & 0.8 & 300 \\
Sleep-EDF & 20 & 0.3 & 0.85 & 300 \\
\bottomrule
\end{tabular}
\label{tab:patch_size}
\end{table}

\subsubsection{Model Architecture}
Our framework is an auto-encoder architecture, which is composed of an encoder and a decoder. Our encoder is based on a deeper Transformer (which is a 6-layer Transformer), and we add CNNs before the Transformer as the local feature extractors. The CNN employed in our experiment is a one-dimensional ResNet-18 architecture \cite{he2016deep}. Like the recent work \cite{he2022masked}, our decoder is a shallower Transformer (which has only two layers), because a lighter decoder can help reduce pre-training time without influencing negatively the performance on down-stream tasks \cite{he2022masked}.

\subsubsection{Data processing}
We use the \textit{fft} function from the \textit{numpy.fft} Python library. For each channel of a time series data, the original transformed sequence is a complex array with the same length. Because of the symmetry, we preserve the former half of the sequence, and store the magnitude and phase of this half sequence. After concatenating three sequences, the new sequence is with the double length of the original time series. When we slice the sequence into patches, we make sure that each patch contains only one type of data (time, magnitude, or phase). We then randomly drop patches in a probability of $r$ (as mentioned in Section \ref{sec:optimization}) in the i-th epoch during pre-training. The patch length and dropping hyper-parameters for each dataset are given in Table \ref{tab:patch_size}.

\subsubsection{Baselines}
We compare our performances with the following self-supervised learning approaches, including the ones designed for time series specifically as well as the ones for general data. 

\begin{itemize}[leftmargin=5mm]
\itemsep0em
\item \textbf{Contrastive Predictive Coding (CPC)} \cite{oord2018representation}: CPC is to learn representations by predicting the future data in the latent space through autoregressive models. It constructs a contrastive loss to maximally preserve the mutual information in the latent space between data in the present time step and data in the future.
\item \textbf{SimCLR} \cite{pmlr-v119-chen20j}: SimCLR is a contrastive learning method firstly designed for image data. It constructs positive pairs and negative pairs by conducting different data augmentation methods. We replace the original encoder architecture with our encoder for a fair comparison. Also, we use the same augmentations as previous work \cite{eldele2021attention} for time series. The same data with two data augmentation methods form positive pairs, and the other pairs are negative pairs. The self-supervised learning task is to maximize the similarity of positive pairs and minimize the similarity of negative pairs in the latent space.
\item \textbf{Temporal and Contextual Contrasting (TS-TCC)} \cite{eldele2021time}: TS-TCC is a self-supervised learning method for time series data. Similar to SimCLR, it transforms the raw time series data into two views by using weak and strong augmentations. The task is a combination of a cross-view prediction task and a contrastive task. The contrastive learning task maximizes the similarity among different views of the same sample while minimizing similarity among views of different samples.
\item \textbf{Temporal Neighborhood Coding (TNC)} \cite{tonekaboni2021unsupervised}: TNC is also a contrastive learning method for time series data. It utilizes the stationary properties of time series data. Similar to SRL, TNC also constructs positive pairs using near segments, and constructs negative pairs using far segments.
\item \textbf{Time Series Transformer (TST)} \cite{zerveas2021transformer}: TST is a self-supervised learning method for time series data by reconstructing the masked parts of original time series data. Different from our method, it neglects the frequency domain and masks the original data by setting the value as 0, which causes a large gap between the pre-training and fine-tuning.
\end{itemize}
      
For fair comparisons, we implement these methods using public code with the same encoder architectures as the original work (except SimCLR). The models are optimized using Adam \cite{kingma2014adam}. The representation dimensions are all set to 256, the same as ours. All experiments are conducted on a server with five NVIDIA GeForce RTX 3090ti GPUs.

\subsubsection{Evaluations}

Linear probing is a popular method to evaluate the self-supervised learning method. However, it restricts the ability of deep learning since it cannot pursue stronger and dataset-specific representations. Hence, we add a two-layer fully connected neural network as the classifier after the Transformer encoder and fine-tune the whole model. The validation set is used to tune the hyper-parameters (such as $beta$ in $\mathcal{L}$, batch size and learning rate) in fine-tuning stage, and the model reaching the highest accuracy on the validation set is saved and evaluated on the test set.

In the self-supervised learning stage, we randomly choose five seeds and get five pre-trained models. In the fine-tuning stage, we also randomly use five seeds to fine-tune each of the five pre-trained models. Thus, we get 25 results per self-supervised learning method per dataset. 

We use ROC-AUC, F1-Score, and Accuracy to evaluate the methods. ROC-AUC is defined as the area surrounded by the Receiver Operator Characteristic curve (ROC curve), the x-axis, and the y-axis. Regarding the ROC curve, it is first computed based on the predicted probability and ground truth of each label directly without a predefined threshold, then defined as the curve of the true positive rate versus the false positive rate at various thresholds ranging from zero to one. Accuracy is calculated for each class, as the ratio of the number of correctly classified samples over the total number of samples. F1 score is the harmonic average of precision (the proportion of true positive cases among the predicted positive cases) and recall (the proportion of positive cases that are correctly identified). As a multi-class classification task, we report average ROC-AUC values across all classes, average Accuracy values across all classes, and macro averaged F1 score. Reported numbers are expressed in percentage values for better reading. 

\subsection{Comparison Results}

We compare our CRT with baseline methods and show the results in Table \ref{tab:baselines}. Our CRT outperforms all baselines in terms of ROC-AUC, F1-score and accuracy. 
We also observe that the results of CRT have a relatively small standard deviation, which implies a more stable performance on downstream tasks. In addition, we notice that TST (another reconstruction-based self-supervised learning method) also obtains a relatively small standard deviation. The reason might be that reconstruction tasks do not require constructing negative and positive pairs, which is difficult and may cause instability. It indicates that reconstruction-based methods may provide a more stable way for self-supervised learning. 
One baseline named Scalable Representation Learning \cite{franceschi2019unsupervised} is not included in our results, as it requires a much longer running time and we failed to produce its results in several days.

\begin{table}[]
    \centering
    \caption{Comparison between baseline methods and our method. The value before $\pm$ is the mean value of multiple-time runs, and the value after $\pm$ is the standard deviation.} 
    \label{tab:baselines}
\resizebox{\columnwidth}{!}{
    \begin{tabular}{c|c|c|c|c}
    \toprule
         Dataset& Method& ROC-AUC  & F1-Score & Accuracy \\
    \midrule
         \multirow{6}{*}{PTB-XL}& CPC & 87.42 $\pm$ 0.34  & 63.17 $\pm$ 1.19 & 85.75 $\pm$ 0.28\\
& SimCLR & 84.50 $\pm$ 1.45 & 59.94 $\pm$ 2.49 & 84.17 $\pm$ 0.80\\
& TSTCC & 87.25 $\pm$ 0.25 & 65.70 $\pm$ 0.71 & 84.66 $\pm$ 1.12\\
& TNC & 86.90 $\pm$ 0.35  & 63.02 $\pm$ 1.13 & 85.35 $\pm$ 0.23\\
& TST & 81.69 $\pm$ 0.39  & 59.36 $\pm$ 8.30 & 81.86 $\pm$ 1.80\\
 & CRT & \textbf{89.22 $\pm$ 0.07} & \textbf{68.43 $\pm$ 0.58} & \textbf{87.81 $\pm$ 0.29}\\

    \midrule
        \multirow{6}{*}{HAR}& CPC & 94.83 $\pm$ 1.15 & 84.42 $\pm$ 2.13 & 83.86 $\pm$ 2.05\\
& SimCLR & 96.53 $\pm$ 0.88 & 86.85 $\pm$ 2.26 & 86.15 $\pm$ 2.37\\
& TSTCC & 97.46 $\pm$ 0.32 & 88.68 $\pm$ 1.43 & 87.95 $\pm$ 1.59\\
& TNC & 94.45 $\pm$ 1.32 & 84.64 $\pm$ 1.96 & 84.01 $\pm$ 1.95\\
& TST & 97.31 $\pm$ 0.39 & 86.17 $\pm$ 1.00 & 85.37 $\pm$ 1.01\\
& CRT & \textbf{98.94 $\pm$ 0.22}  & \textbf{90.51 $\pm$ 0.77} & \textbf{90.09 $\pm$ 0.75}\\
    \midrule
    
\multirow{6}{*}{Sleep-EDF}& CPC & 91.92 $\pm$ 1.62  & 39.74 $\pm$ 3.35 & 88.70 $\pm$ 1.35\\
& SimCLR & 91.97 $\pm$ 3.22  & 41.78 $\pm$ 3.13 & 87.86 $\pm$ 1.77\\
& TSTCC & 93.57 $\pm$ 1.83 & 39.09 $\pm$ 4.50 & 86.06 $\pm$ 3.19\\
& TNC & 91.48 $\pm$ 3.51  & 37.89 $\pm$ 5.13 & 86.97 $\pm$ 3.48\\
& TST & 93.31 $\pm$ 2.29  & 42.58 $\pm$ 2.48 & 88.83 $\pm$ 0.84\\
& CRT & \textbf{94.74 $\pm$ 1.09} & \textbf{44.38 $\pm$ 1.14} & \textbf{90.12 $\pm$ 0.57}\\
\bottomrule
    \end{tabular}
    
}
\end{table}

\subsection{Analysis of Cross-Domain Reconstruction}

More ablation studies and comparison experiments are conducted to further verify the effectiveness of our method.

\subsubsection{Dropping is Better than Masking}

\begin{figure}
    \centering
    \includegraphics[width=\linewidth]{ 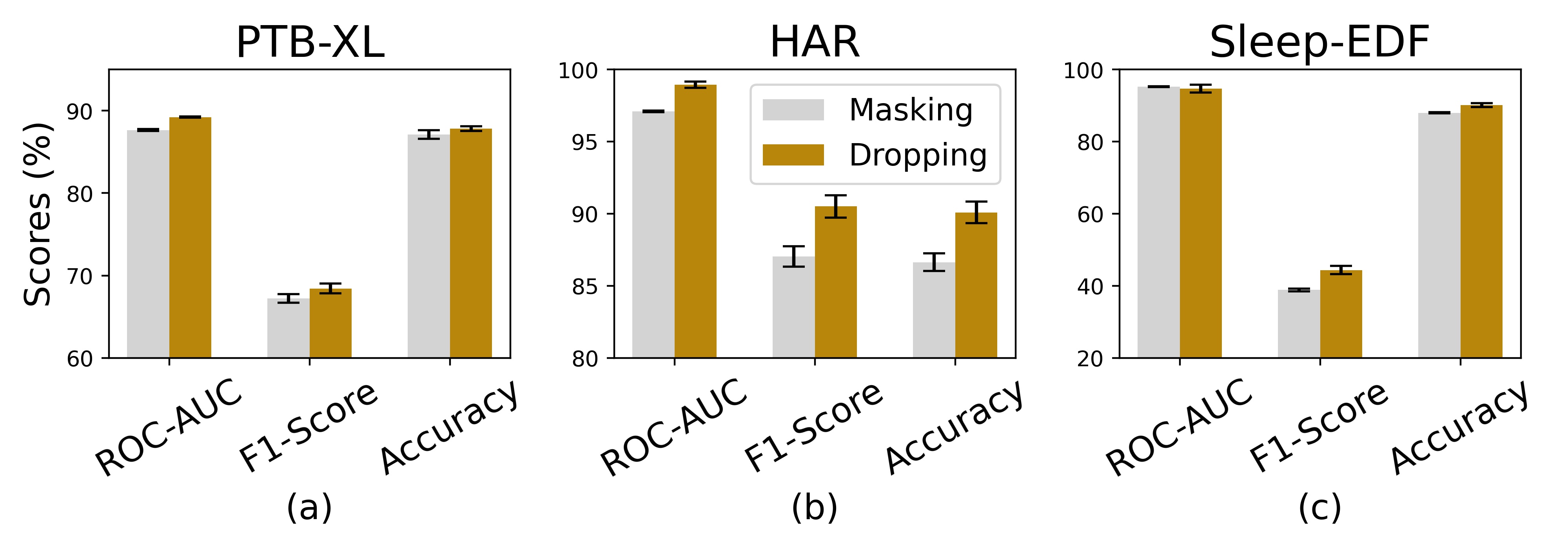}
    \caption{The performance on three datasets when data is dropped or masked during the pre-training stage.}
    \label{fig:drop_mask}
\end{figure}

To verify that dropping can better alleviate the gap between pre-training and fine-tuning than masking, we compare the results on downstream tasks under the two setups. From Figure \ref{fig:drop_mask}, we see dropping always leads to better downstream performances, except for the Sleep-EDF dataset, where dropping yields a slightly lower ROC-AUC but outperforms masking for other metrics. This is because zeros brought by masking may create wrong patterns for time series data, and dropping can tackle this problem. Our dropping approach can keep the shape of the original time series, and reduce the gap between the pre-training and fine-tuning phases caused by the masking process.

\subsubsection{Adding Phase Helps Frequency Learning}

\begin{figure}
    \centering
    \includegraphics[width=\linewidth]{ 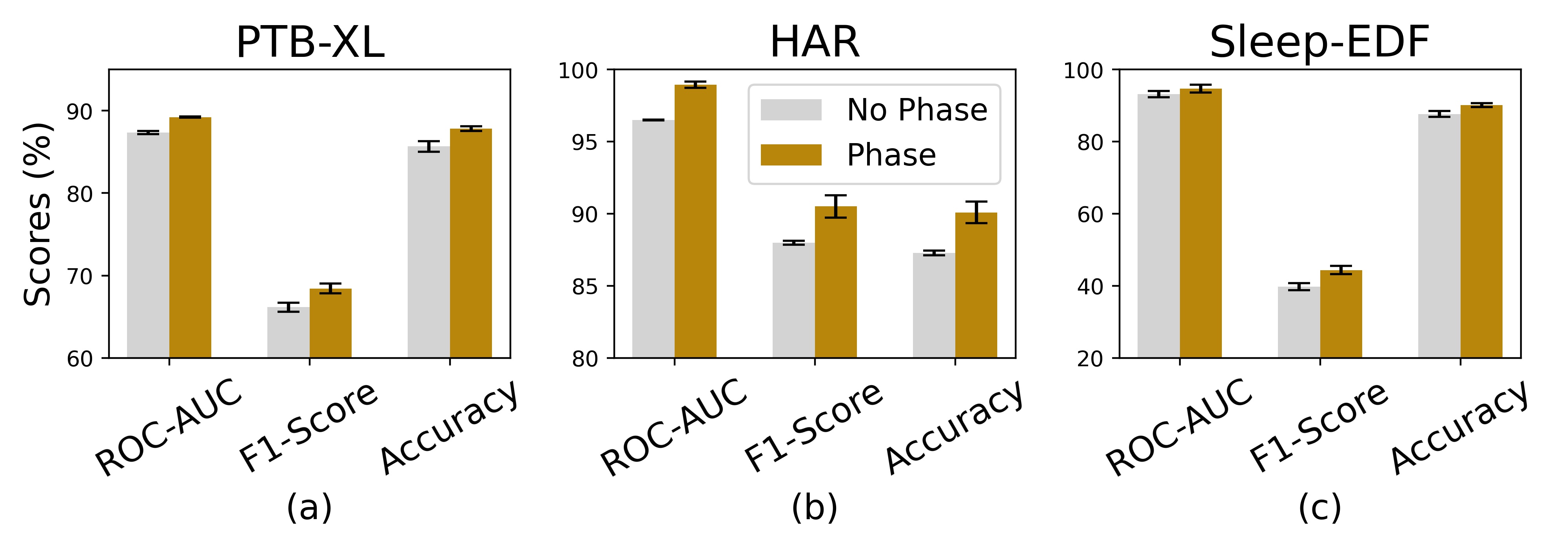}
    \caption{The performance on three datasets with and without phase.}
    \label{fig:phase_nophase}
\end{figure}

We also conduct a simple ablation study to support that phase can provide supplementary spectral information. We evaluate the model pre-trained and fine-tuned with and without phase data under the same experimental setups as our CRT. The results on three datasets with and without phase data are shown in Figure \ref{fig:phase_nophase}. We can see that adding the phase can help improve the final performance in terms of three metrics. This supports our view that magnitude is not informative enough to represent the frequency domain, and phase can complement magnitude.

\subsubsection{Cross-Domain is Better than Single-Domain}\label{sec:cross_single}

\begin{table}[]
    \centering
    \caption{The comparison of cross-domain and single-domain. Time represents time-to-time reconstruction and Freq represents frequency-to-frequency reconstruction. }
    \resizebox{\columnwidth}{!}{
    \begin{tabular}{c|c|c|c|c}
    \toprule
    Dataset & Method & ROC-AUC & F1-Score & Accuracy \\
    \midrule
      \multirow{3}{*}{PTB-XL}  & Time & 88.08 $\pm$ 0.25  & 67.45 $\pm$ 0.55 & 85.67 $\pm$ 0.59\\
    & Freq & 77.58 $\pm$ 0.43 & 47.86 $\pm$ 1.96 & 78.28 $\pm$ 0.91\\
    & CRT & \textbf{89.22 $\pm$ 0.07} & \textbf{68.43 $\pm$ 0.58} & \textbf{87.81 $\pm$ 0.29}\\

    \midrule
    \multirow{3}{*}{HAR}  & Time & 96.46 $\pm$ 0.13 & 87.62 $\pm$ 1.00 & 86.84 $\pm$ 1.07\\
    & Freq & 94.61 $\pm$ 0.14 & 76.96 $\pm$ 1.38 & 76.50 $\pm$ 1.44\\
    & CRT & \textbf{98.94 $\pm$ 0.22}  & \textbf{90.51 $\pm$ 0.77} & \textbf{90.09 $\pm$ 0.75}\\

    \midrule
    \multirow{3}{*}{Sleep-EDF}  & Time & 94.64 $\pm$ 0.55 & 40.13 $\pm$ 1.30 & 88.25 $\pm$ 0.59\\
    & Freq& 91.00 $\pm$ 0.76 & 38.07 $\pm$ 1.88 & 84.37 $\pm$ 1.33\\
    & CRT & \textbf{94.74 $\pm$ 1.09}  & \textbf{44.38 $\pm$ 1.14} & \textbf{90.12 $\pm$ 0.57}\\
    \bottomrule
    \end{tabular}
    }
    \label{tab:single_multiple}
\end{table}

\begin{figure}
    \centering
    \subfloat[]{\includegraphics[width=0.6\linewidth]{ 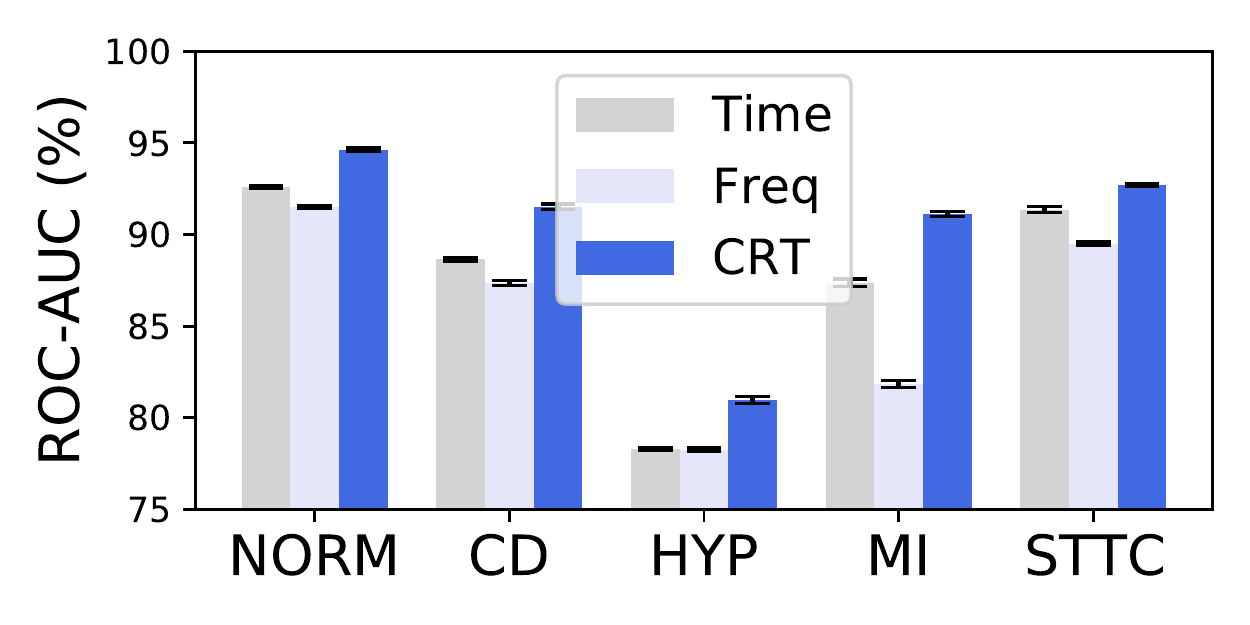} \label{fig:byclass_ptbxl1} }
    \subfloat[]{
    \includegraphics[width=0.4\linewidth]{ 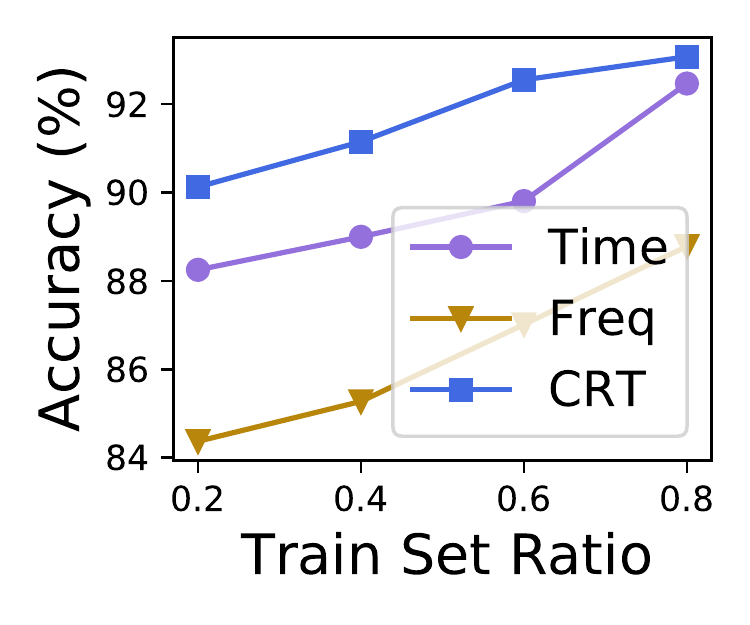}\label{fig:byclass_ptbxl2}}
    \caption{The performance when using time domain, frequency domain, and both domains. \textbf{(a)}: ROC-AUC of all 5 classes on PTB-XL. \textbf{(b)}: Accuracy on Sleep-EDF when the ratio of the training set used for fine-tuning is 0.2, 0.4, 0.6, and 0.8.}
    \label{fig:byclass_ptbxl}
\end{figure}

We compare our method with the trivial single-domain reconstruction tasks. After conducting time-domain and frequency-domain (phase and magnitude) reconstruction tasks respectively, we fine-tune the models using the corresponding domain's data. As shown in Table \ref{tab:single_multiple}, CRT performs best in terms of all 3 metrics. The ROC-AUCs of all five classes on PTB-XL datasets are shown in Figure \ref{fig:byclass_ptbxl1}. We can see that cross-domain representations help improve performance on all classes. This is because some abnormalities of ECGs can be more easily observed from the perspective of the frequency domain. Figure \ref{fig:byclass_ptbxl2} shows the performance on the Sleep-EDF dataset under different training sizes. We give the results when 20\%, 40\%, 60\%, and 80\% of the training set is used for fine-tuning. The results show that by combining two domains, our cross-domain representations can yield higher accuracy under different ratios.

All results support our hypothesis that learning from two domains can yield more informative representations than from a single domain. This is because some patterns of the frequency domain can complement the temporal patterns. Also, the cross-domain Transformer encoder can fuse different types of features, and extract useful and complementary patterns from both domains. 

In addition, we notice that in all three datasets, using the time domain yields better performance compared with using the frequency domain. This may be because the patterns of the time domain are more direct for neural networks to discover, and the three classification tasks are strongly related to these patterns. However, adding the frequency domain can supplement some patterns, leading to better performance.

\subsubsection{Exploring Cross-Domain Learning}

\begin{figure}
\centering

\subfloat[]{
\includegraphics[width=0.99\linewidth]{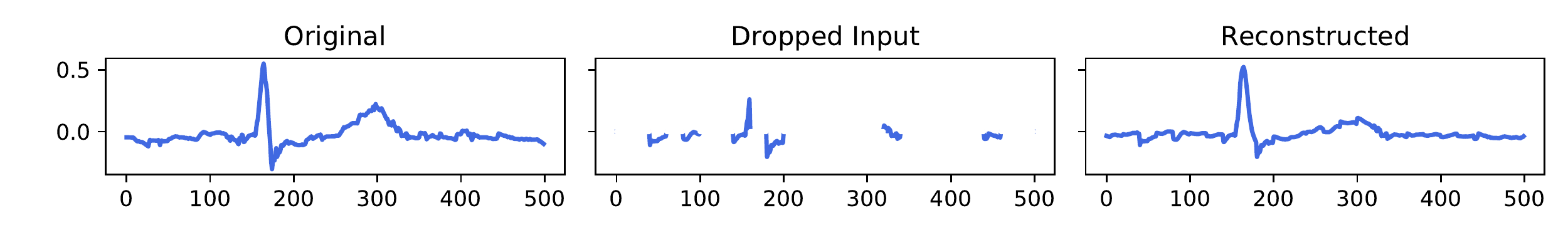}
}\label{fig:case_reconstruction1}
\subfloat[]{
\includegraphics[width=0.99\linewidth]{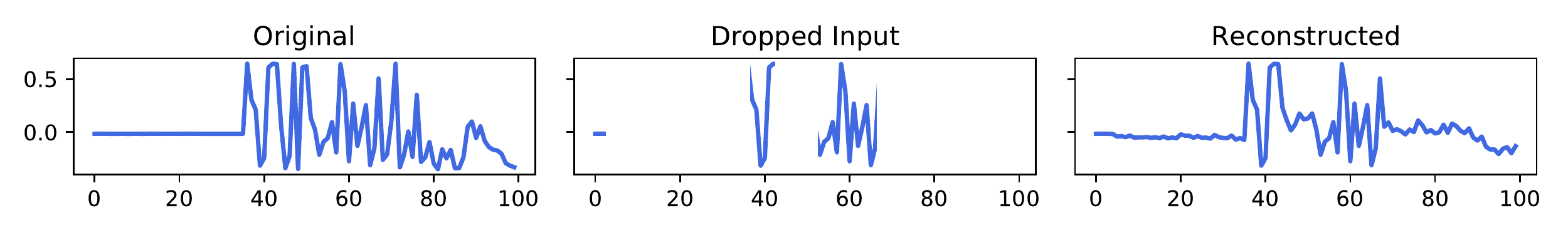}
}\label{fig:case_reconstruction2}
\subfloat[]{
\includegraphics[width=0.99\linewidth]{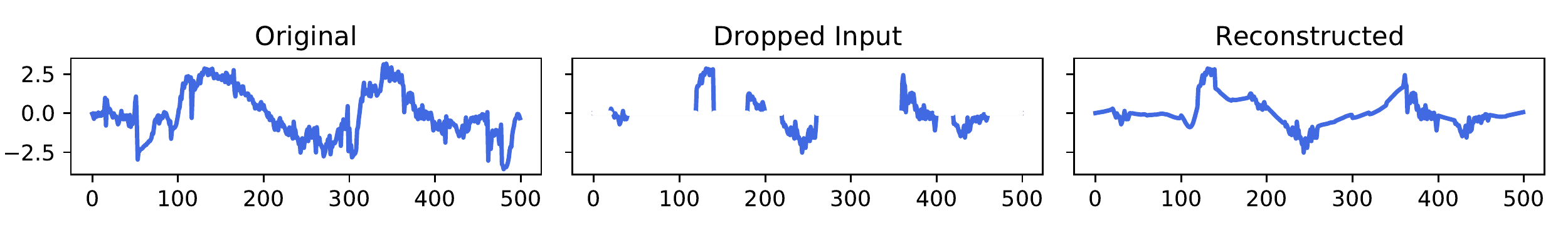}
}\label{fig:case_reconstruction3}
\caption{Three reconstructed cases (the ``Reconstructed'' row) from PTB-XL (a), HAR (b) and Sleep-EDF (c). ``Original'' represents the original time series, and ``Dropped input'' represents the time series after being dropped.}\label{fig:case_reconstruction}
\end{figure}

\begin{table}[]
    \centering
    \caption{Comparisons of three kinds of cross-domain reconstruction. T2F represents Time-to-Frequency and F2T represents Frequency-to-Time.}
    \resizebox{\columnwidth}{!}{
    \begin{tabular}{c|c|c|c|c}
    \toprule
    Dataset & Method & ROC-AUC & F1-Score & Accuracy \\
    \midrule
      \multirow{3}{*}{PTB-XL}  & T2F & 84.14 $\pm$ 0.34 & 59.02 $\pm$ 2.04 & 83.58 $\pm$ 0.78 \\
        & F2T & 83.72 $\pm$ 0.49 & 59.19 $\pm$ 1.95 & 83.42 $\pm$ 0.70 \\
        & CRT & \textbf{89.22 $\pm$ 0.07} & \textbf{68.43 $\pm$ 0.58} & \textbf{87.81 $\pm$ 0.29}\\
    \midrule
    \multirow{3}{*}{HAR} 
    & T2F & 98.18 $\pm$ 0.58 & 89.23 $\pm$ 1.64 & 88.84 $\pm$ 1.70\\
    & F2T & 97.93 $\pm$ 0.83 & 88.90 $\pm$ 1.97 & 88.53 $\pm$ 1.98 \\
    & CRT & \textbf{98.94 $\pm$ 0.22}  & \textbf{90.51 $\pm$ 0.77} & \textbf{90.09 $\pm$ 0.75}\\
    \midrule
    \multirow{3}{*}{Sleep-EDF}  & T2F & 94.36 $\pm$ 1.66 & 40.41 $\pm$ 3.65 & 89.09 $\pm$ 2.62 \\
     & F2T & 92.88 $\pm$ 1.53 & 41.67 $\pm$ 1.93 & 89.46 $\pm$ 1.00 \\
     & CRT & \textbf{94.74 $\pm$ 1.09}  & \textbf{44.38 $\pm$ 1.14} & \textbf{90.12 $\pm$ 0.57}\\
    \bottomrule
    \end{tabular}
    }
    \label{tab:cross_domain}
\end{table}

\begin{figure}
    \centering
    \subfloat[]{\includegraphics[width=0.55\linewidth]{ 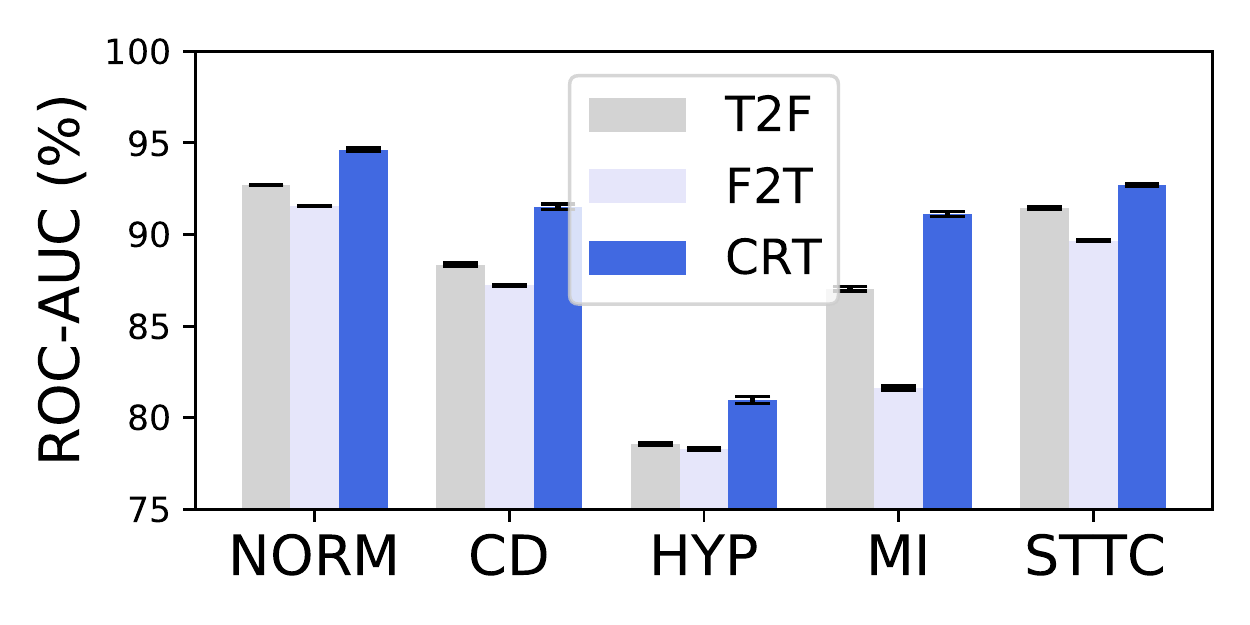} \label{fig:byclass_ptbxl_f2tt2f1}}
    \subfloat[]{\includegraphics[width=0.35\linewidth]{ 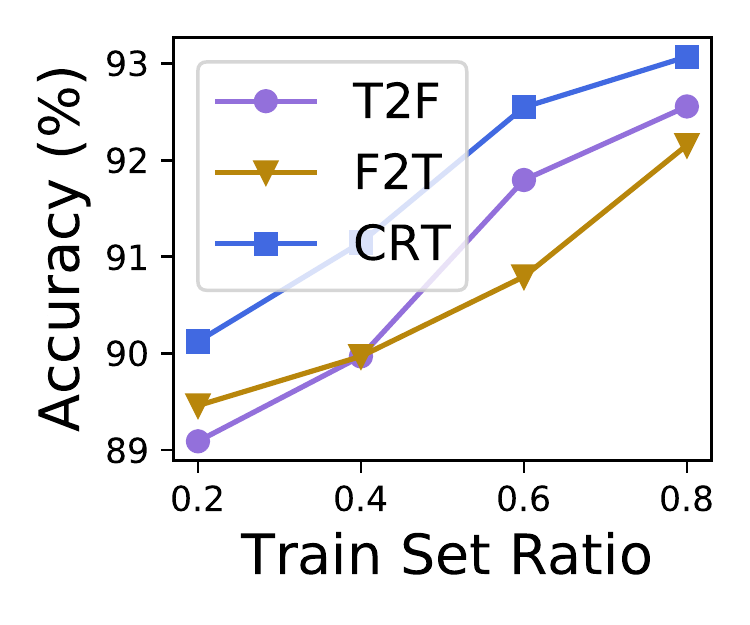} \label{fig:byclass_ptbxl_f2tt2f2}}

    \caption{The performance of three cross-domain reconstruction methods. \textbf{(a)}: ROC-AUC of all 5 classes on PTB-XL. \textbf{(b)}: Accuracy on Sleep-EDF when the ratio of the training set used for fine-tuning is 0.2, 0.4, 0.6 and 0.8.}
    \label{fig:byclass_ptbxl_f2tt2f}
\end{figure}

We explore cross-domain learning with two experiments:
\begin{itemize}[leftmargin=5mm]
\itemsep0em
\item \textbf{Time-to-Frenquency Reconstruction (T2F)}: with only the time-domain part of $\mathbf{E}$ fed into the Transformer encoder, we use the encoded features to reconstruct the dropped original frequency-domain data (including phase and magnitude).
\item \textbf{Frequency-to-Time Reconstruction (F2T)}: contrary to T2F, we input the frequency-domain part of $\mathbf{E}$ and reconstruct the dropped time-domain data.
\end{itemize}
We used the same inputs for fine-tuning to only evaluate the cross-reconstruction tasks.

As shown in Table \ref{tab:cross_domain}, our complete cross-domain solution outperforms the other two methods on three datasets. We also give the results of all five classes on the PTB-XL dataset (Figure \ref{fig:byclass_ptbxl_f2tt2f1}), and the accuracy on the Sleep-EDF dataset under four training set ratios (Figure \ref{fig:byclass_ptbxl_f2tt2f2}). From Figure \ref{fig:byclass_ptbxl_f2tt2f1}, we can see CRT outperforms others on all classes, especially on ``MI'' (11.6\% higher than F2T and 4.7\% higher than T2F). In Figure \ref{fig:byclass_ptbxl_f2tt2f2}, CRT also performs better, and T2F performs similarly to F2T. 

Our CRT adopts Transformer as part of our encoder to fuse the embeddings from different domains. The representations obtained from the Transformer encoder contain information of both time domain and frequency domain, which benefits the reconstruction as well as the downstream tasks. T2F and F2T, on the other hand, only capture a single-direction relationship rather than the mutual relationship. It is intuitive that CRT can produce more semantic representations, and the experiment results well prove it.

\subsubsection{CL and IDC Help Learning Cross-Domain Representations}
\begin{figure}
    \centering
    \includegraphics[width=0.9\linewidth]{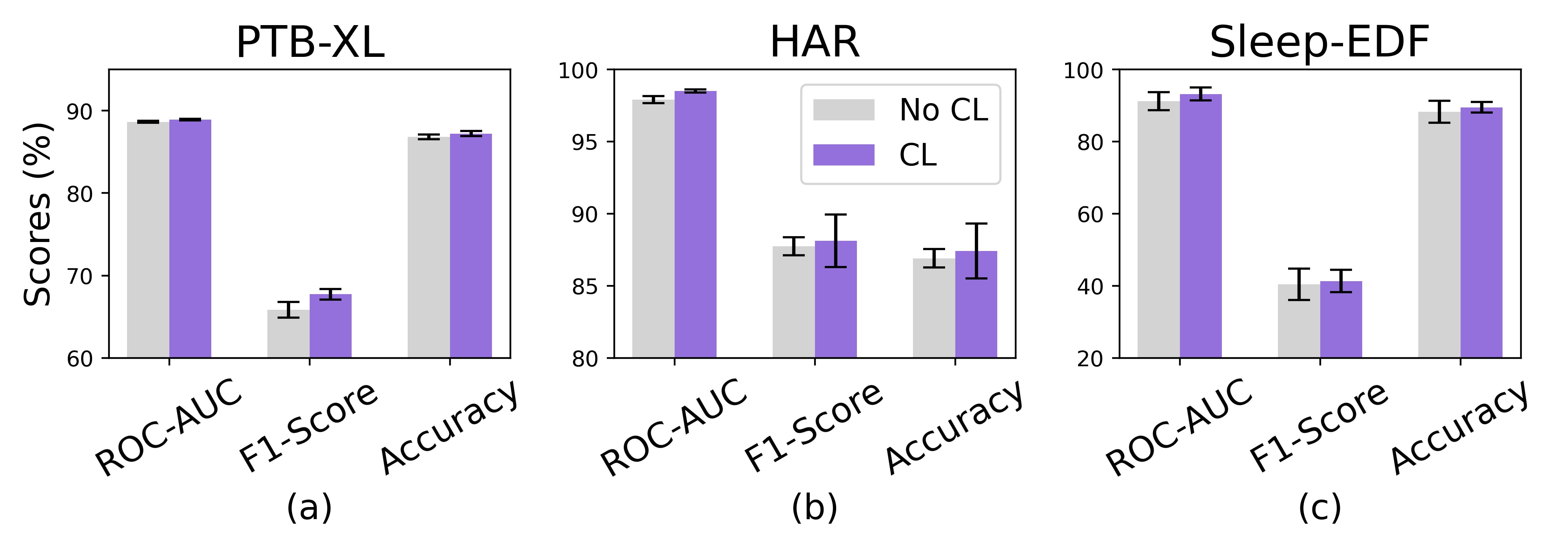}
    \caption{The performance on three datasets with and without CL.}
    \label{fig:cl_nocl}
\end{figure}

\begin{figure}
    \centering
    \includegraphics[width=0.9\linewidth]{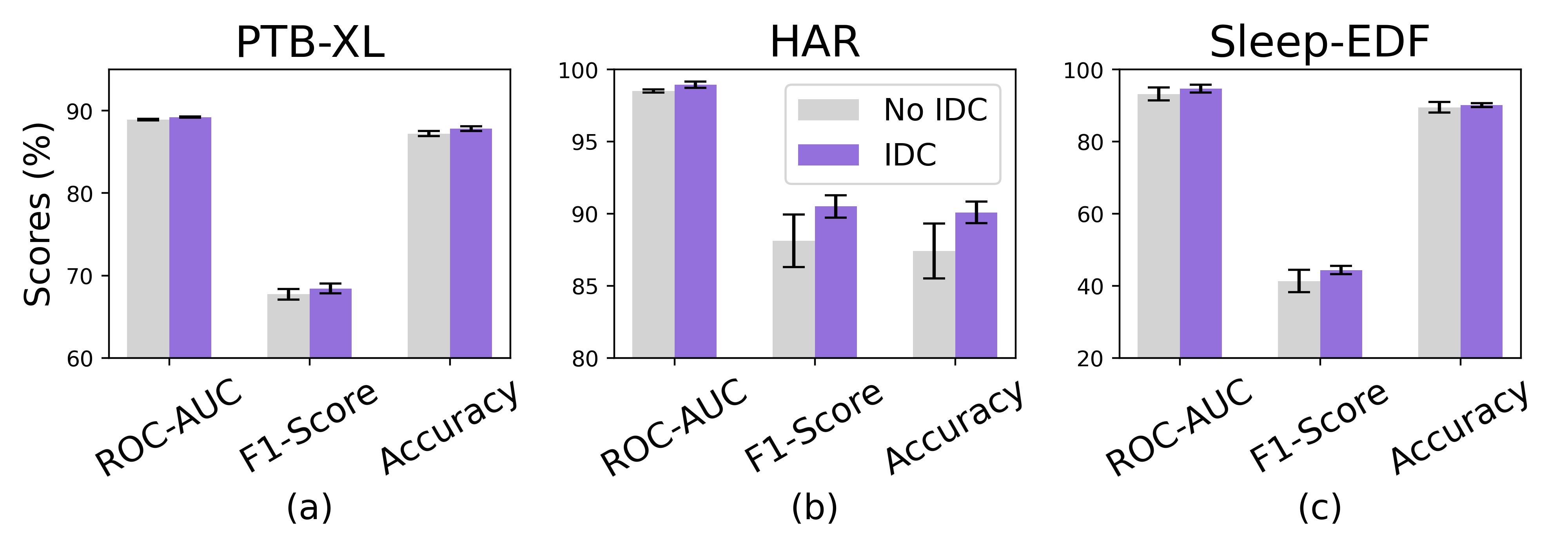}
    \caption{The performance on three datasets with and without IDC.}
    \label{fig:ib_noib}
\end{figure}

We conduct another ablation study to verify the effectiveness of CL and IDC, and show the results in Figure \ref{fig:cl_nocl} and Figure \ref{fig:ib_noib}.

As we expect, these two modules facilitate learning cross-domain representations. CL helps the reconstruction tasks and improves the performance on downstream tasks. CL gives a progressively increasing dropping ratio rather than a constant large dropping ratio, which helps better reconstruct the original data. As a regularization, IDC minimizes the mutual information between different samples which helps Transformer extract more discriminable patterns. It is worth noting that IDC is not a contrastive loss, since it does not minimize the distance of any pair of samples. It only minimizes common information of all time series.

In Figure \ref{fig:case_reconstruction}, we illustrate three reconstructed cases with a 0.7 dropping ratio. We see that the model almost reconstructs the trend of dropped inputs with such a high dropping ratio, especially for the ECG case from the PTB-XL dataset. The reason for the well-performed ECG reconstruction might be because that ECG is periodic and there are abundant similar patterns in the time series.
Such amazing reconstruction performance also proves that our cross-domain representations are highly informative and sample-specific.

\section{Conclusion}
In this work, we propose Cross Reconstruction Transformer (CRT), a cross-domain reconstruction framework based on Transformer for self-supervised representations learning of time series data. We note that the existing self-supervised learning methods neglect to utilize the spectral information and temporal-spectral correlations of time series. Moreover, existing ``masking''-based methods for reconstruction tasks tend to significantly change the original pattern of time series, which would lead to the distribution shifts between pre-training and fine-tuning processes. To tackle these problems in a unified way, we propose our cross-domain reconstruction framework which outperforms existing methods on three widely-used datasets. In addition, we conduct various experiments to verify the effectiveness of each component of our framework. For future work, we would adapt our framework to the out-of-distribution scenarios and address more challenging problems \cite{yang2022diffusion}.

\section*{Acknowledgement}
This work was supported by the National Natural Science Foundation of China (No.62102008).

\bibliography{CRT}
\bibliographystyle{IEEEtran}

\end{document}